\documentclass{article}

\usepackage[numbers]{natbib}

\usepackage[final]{neurips_2020}

\usepackage[utf8]{inputenc} 
\usepackage[T1]{fontenc}    
\usepackage{hyperref}       
\usepackage{url}            
\usepackage{booktabs}       
\usepackage{amsfonts}       
\usepackage{nicefrac}       
\usepackage{microtype}      

\usepackage{graphicx}
\usepackage{multirow}
\usepackage{amsmath}
\usepackage{amssymb}
\usepackage{subcaption}
\usepackage{xcolor}
\usepackage{booktabs}
\usepackage{floatrow}
\newfloatcommand{capbtabbox}{table}[][\FBwidth]
\usepackage{blindtext}

\newcommand{\wc}[1]{\textbf{TODO: {#1}}}

\newcommand{\emql}{\textsf{EmQL}}
\newcommand{\emqlf}{\mbox{\emql{}}-\mbox{filter}}
\newcommand{\emqls}{\mbox{\emql{}}-\mbox{sketch}}
\newcommand{\abs}[1]{\lvert#1\rvert}

\newcommand{\R}{{\rm I\!R}}

\newcommand{\vek}[1]{\textbf{#1}}

\newcommand{\vekA}{\vek{v}_A}
\newcommand{\vekB}{\vek{v}_B}

\newtheorem{theorem}{Theorem}

\title{Faithful Embeddings for Knowledge Base Queries}

\author{
  {Haitian Sun \qquad Andrew O. Arnold\thanks{Work done while at Google Research.} \qquad Tania Bedrax-Weiss}\\
  {\bf Fernando Pereira \qquad William W. Cohen}\\
  Google Research \\
  \texttt{\{haitiansun,tbedrax,pereira,wcohen\}@google.com}\\
  AWS AI \\
  \texttt{anarnld@amazon.com}\\  
}
\begin{document}
\maketitle

\begin{abstract}
The deductive closure of an ideal knowledge base (KB) contains exactly the logical queries that the KB can answer. However, in practice KBs are both incomplete and over-specified, failing to answer some queries that have real-world answers. \emph{Query embedding} (QE) techniques have been recently proposed where KB entities and KB queries are represented jointly in an embedding space, supporting relaxation and generalization in KB inference. However, experiments in this paper show that QE systems may disagree with deductive reasoning on answers that do not require generalization or relaxation.  We address this problem with a novel QE method that is more faithful to deductive reasoning, and show that this leads to better performance on complex queries to incomplete KBs. Finally we show that inserting this new QE module into a neural question-answering system leads to substantial improvements over the state-of-the-art. \footnote{Code available at \url{https://github.com/google-research/language}}
\end{abstract}

\section{Introduction}
The deductive closure of an ideal knowledge base (KB) contains exactly the logical queries that the KB can answer. However, in practice KBs are both incomplete and over-specified, failing to answer queries that have actual real-world answers. \emph{Query embedding} (QE) methods extend logical queries to incomplete KBs by representing KB entities and KB
queries in a joint embedding space, supporting relaxation and generalization in KB inference~\citep{guu2015traversing,hamilton2018embedding,emptysparql2018,ren2020query2box}. For instance, graph query embedding (GQE)~\citep{hamilton2018embedding} encodes a query $q$ and entities $x$ as vectors such that cosine
distance represents $x$'s score as a possible answer to $q$. In QE, the embedding for a query $q$ is typically built compositionally; in particular, the embedding for $q=q_1\wedge{}q_2$ is computed from the embeddings for $q_1$ and $q_2$.   In past work, QE has been
useful for answering overconstrained logical queries~\cite{emptysparql2018} and querying {incomplete} KBs~\cite{guu2015traversing,hamilton2018embedding,ren2020query2box}.

\begin{figure}
\includegraphics[width=0.3\textwidth]{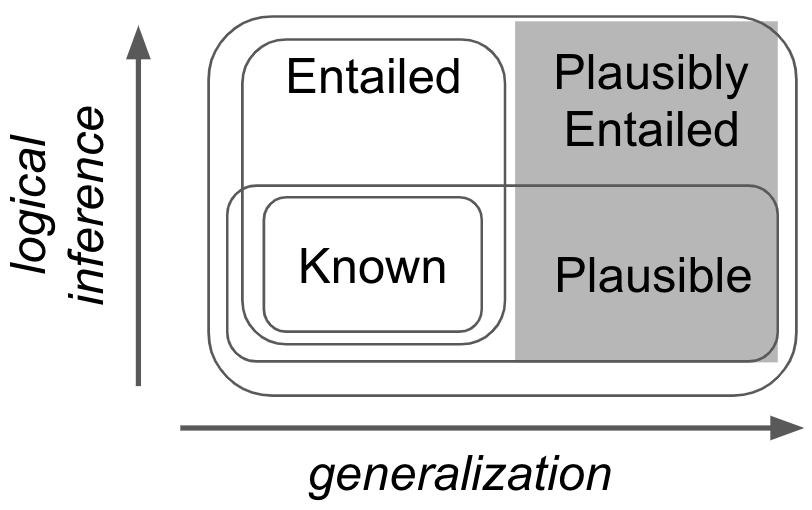}~~~~\begin{minipage}[b]{0.65\textwidth}
  KB embedding (KBE) methods generalize from \emph{known} KG
  facts to \emph{plausible} ones, and logical inference
  computes answers to compositional queries that are
  \emph{entailed} by known facts.  Query embedding (QE) combines both
  of these tools for extending a set of known facts, by finding
  answers to a query that are \emph{plausibly entailed} by known
  facts.\\
\end{minipage}
\vspace{-6pt}
\caption{Overview of differences between KBE and QE.  Shaded area indicates the
  kinds of test cases used in prior studies of QE.} \label{fig:qe-vs-kbe}
\end{figure}

Figure~\ref{fig:qe-vs-kbe} summarizes the relationship between traditional KB
embedding (KBE), query embedding (QE), and logical inference. Traditional
logical inference enables a system to find deductively \emph{entailed} answers to queries; KBE approaches allow a system to \emph{generalize} from explicitly-stored KB tuples to similar tuples; and 
QE methods combine both of these ideas, providing a soft form of
logical entailment that generalizes.

We say that a QE system is \emph{logically faithful} if it behaves
similarly to a traditional logical inference system with respect to entailed answers.  In this paper, we present experiments illustrating that QE systems are often \emph{not} faithful: in particular, experiments with the state-of-the-art QE system Query2Box~\cite{ren2020query2box} show that
it performs quite poorly in finding logically-entailed answers.  
We conjecture this is because models that generalize well
do not have the capacity to model all the information in a large KB
accurately, unless embeddings are impractically large.
We thus propose two novel methods for improving faithfulness while
preserving the ability to generalize.  First, we implement some
logical operations using neural retrieval over a KB of embedded
triples, rather than with geometric operations in embedding
space, thus adding a non-parametric component to QE.
Second, we employ a randomized data structure called a count-min sketch to propagate
scores of logically-entailed answers.  We show that this combination
leads to a QE method, called \emql{} (Embedding Query
  Language) which is differentiable, compact, scalable, and (with
high probability) faithful.  Furthermore, strategically removing the sketch in parts of the QE system allows it to generalize very effectively. 

We show that \emql{} performs dramatically better than Query2Box on
logically-entailed answers, and also improves substantially on complex
queries involving generalization.  Finally we show that
inserting \emql{} into a natural language KB question-answering (KBQA)
system leads to substantial improvements over the experimental
state-of-the-art for two widely-used benchmarks, MetaQA~\cite{zhang2017variational} and WebQuestionsSP~\cite{webqsp}.

The main contributions of this work are: (1) a new QE scheme with expressive set and
relational operators, including those from
previous QE schemes (set intersection, union, and relation
following) plus a ``relational filtering'' operation; (2) a new analysis of QE methods showing that previous methods
are not faithful, failing to find entities
logically entailed as answers; (3) the first application of
QE as a module in a KBQA system; and (4) evidence that this module
leads to substantial gains over the prior state-of-the-art on two
widely-used benchmarks, thanks to its superior faithfulness.

\section{Related work} \label{sec:related}

\textbf{KBE and reasoning.}  There are many KB embedding (KBE)
methods, surveyed in~\cite{wang2017knowledge}.  Typically KBE methods
generalize a KB by learning a model that scores the plausibility of a
potential KB triple $r(x,y)$,  where $r$ is a KB relation, $x$ is a head
(aka subject) entity, and $y$ is a tail (aka object) entity.  In
nearly all KBE models, the triple-scoring model assumes that every
entity $x$ is represented by a vector $\vek{v}_x$.

Traditional query languages for symbolic KBs do support testing
whether a triple is present in a KB, but also allow expressive
compositional queries, often queries that return sets of entities.
Several previous works also propose representing \emph{sets} of
entities with embeddings~\cite{vilnis2014word,vilnis2018probabilistic,zaheer2017deep}; box embeddings~\cite{vilnis2018probabilistic}, for instance, represent sets with axis-parallel hyperrectangles.


Many KBE models also support \emph{relation projection}, sometimes
also called \emph{relation following}.  \emph{Relation following}
\cite{cohen2020scalable} maps a set of entities $X$ and a set of
relations $R$ to a set of entities related to something in $X$ via some
relation in $R$: here we use the notation \( X.\textit{follow}(R)
\equiv \{ y ~|~ \exists x\in{}X, r\in{}R : r(x,y)\} \).  Many KBEs
naturally allow computation of some soft version of relation
following, perhaps restricted to singleton sets.\footnote{E.g.,
  translational embedding schemes like
  TransE~\cite{bordes2013translating} would estimate the embedding for
  $y$ as $\hat{\vek{e}}_y = \vek{e}_x + \vek{e}_r$, where $\vek{e}_x$,
  and $\vek{e}_r$ are vectors embedding entity $x$ and relation $r$
  respectively.  Several other
  methods~\cite{guu2015traversing,liu2017analogical} estimate
  $\hat{\vek{e}}_y = \vek{e}_x \vek{M}_r$ where $\vek{M}_r$ is a matrix
  representing $r$.}  However, most KBE methods give poor results when
relation following operations are composed~\cite{guu2015traversing},
as in computing $X.\textit{follow}(R_1).follow(R_2)$.  To address
this, some KBE systems explicitly learn to follow a path (aka chain) of
relations~\cite{guu2015traversing,lin2015modeling,das2016chains}.

Extending this idea, the graph-query embedding (GQE)
method~\cite{hamilton2018embedding} defined a query language
containing both {relation following} and set intersection. In GQE
inputs and output to the relation following operation are
entity sets, defined by cosine-distance proximity to a central vector.
More recently, the Query2Box \cite{ren2020query2box} method varied GQE
by adopting a box embedding for sets, and also extended GQE by
including a set union operator.  In Query2Box unions are implemented
by rewriting queries into a normal form where unions are used only as
the outermost operation, and then representing set unions as unions of
the associated boxes.

Quantum logic~\cite{svozil1998quantum} is another neural
representation scheme, which might be considered a query language.  It
does not include relation following, but is closed under intersection
and negation, and approximately closed under union (via computation of
an upper bound on the union set.)

Other studies~\cite{demeester2016lifted,rastogi2017training} use logical constraints such as transitivity or implication to improve embeddings.  Here we go in the opposite direction, from KBE to reasoning, answering compositional queries in an
embedded KB that is formed in the absence of prior knowledge about relations.

\textbf{Sparse-matrix neural reasoning.}  An alternative to
representing entity sets with embeddings is to represent sets with
``$k$-hot'' vectors. 
Set operations are easily performed on $k$-hot vectors\footnote{If
   $\vekA,\vekB$ are $k$-hot vectors for sets $A,B$, then $\vekA
  + \vekB$ encodes $A \cup B$ and $\vekA \odot \vekB$ encodes $A
  \cap B$.}  and relation following can be implemented as matrix
multiplication~\cite{nql2019}.  Such ``localist'' representations can
exactly emulate logical, hence faithful, reasoning systems. However, they do not offer a direct way to generalize because entities are just (represented by) array indices.



\section{Faithful queries on an embedded KB} \label{sec:method}

\textbf{Background and notation.}  The query language
\emql{} operates on weighted sets of entities.  Let $U$ be the set of all
entities in a KB. A weighted set $X \subseteq U$ is \emph{canonically
  encoded} as a $k$-hot vector $\vek{v}_X \in \R^N$, where $N=|U|$ and
$\vek{v}_X[i]$ holds the non-negative real \emph{weight} of element
$i$ in $X$.
However the $k$-hot encoding is very inefficient if $N$ is large, which we address later. \emql{} relies on a learned embedding $\vek{e}_i\in\R^d$ for each entity $i$, which together form the matrix $\vek{E}\in\R^{d\times N}$ of entity embeddings. 
A weighted set $X$ will be represented by a pair consisting of (1) a dense vector derived from its entity embeddings $\{\vek{e}_i\}$, $i\in X$, plus an efficient sparse representation of the weights $\vek{v}_X[i]$.

In addition to (weighted) set intersection, union, and difference, which are common to many KBE models, 
\emql{} implements two operators for relational reasoning: \emph{relation following} and \emph{relational filtering}.  
\emql{} also supports a limited form of set difference (see Supplemental Material C.)
In this section, we will start by discussing how to encode and decode sets with \emql{} representations, and then discuss the operators in \emql{} for relational reasoning.

\textbf{Representing sets.}  We would like to represent entity sets with a
scheme that supports generalization, but also allows for precisely
encoding weights of sets that are defined by compositional logic-like operations.  
Our representation will assume that sets are of limited cardinality,
and contain ``similar'' entities (as defined below).

We represent a set $X$ with the pair $(\textbf{a}_X, \vek{b}_X)$,
\(\vek{a}_X = \sum_{i} \vek{v}_X[i] ~\vek{e}_i \), \(\vek{b}_X
=\vek{S}_H(\vek{v}_X) \) where $\textbf{a}_X$ is the weighted 
centroid of elements of X that identifies the
general region containing elements of $X$, and $\vek{b}_X$ is an
optional count-min sketch~\cite{cormode2005improved}, which encodes
additional information on the weights of elements of $X$.
Count-min sketches \cite{cormode2005improved}
are a widely used randomized data structure that can approximate the vector $\vek{v}_X$ with limited storage. Supplemental Material B summarizes more technical details, but we summarize count-min sketches below. Our analysis largely follows \cite{daniely2016sketching}.

Let $h$ be a hash function mapping $\{1,\ldots,N\}$ to a smaller range
of integers $\{1,\ldots,N_W\}$, where $N_W \ll N$.  The
\emph{primitive sketch of $\vek{v}_X$ under $h$}, written
$\vek{s}_h(\vek{v}_X)$, is a vector such that
$$ \vek{s}_h(\vek{v}_X)[j] = \sum_{i:h(i)=j} \vek{v}_X[i]
$$ Algorithmically, this vector could be formed by starting with an
all-zero's vector of length $N_W$, then looping over every pair
$(i,w_i)$ where $w_i=\vek{v}_X[i]$ and incrementing each $\vek{s}_h[j]$ by $w_i$. 

A primitive sketch $\vek{s}_h$ contains some information about
$\vek{v}_X$: to look up the value $\vek{v}_X[i]$, we could look up
$\vek{s}_h[h(i)]$, and this will have the correct value if no other set
element $i'$ hashed to the same location.  We can improve this by
using multiple hash functions. Specifically, let
$H=\{h_1,\ldots,h_{N_D}\}$ be a list of $N_D$ hash functions mapping
$\{1,\ldots,N\}$ to the smaller range of integers $\{1,\ldots,N_W\}$.
The \emph{count-min sketch $\vek{S}_H(\vek{v}_X)$ for a $\vek{v}_X$ under $H$} is a matrix such that each row $j$ is the primitive
sketch of $\vek{v}_X$ under $h_j$.  This sketch is an $N_W \times N_D$
matrix, where $N_W$ is called the sketch \emph{width} and $N_D$ is called the
sketch \emph{depth}.

Let $\vek{b}_X = \vek{S}_H(\vek{v}_X)$ be the count-min sketch for $X$.  To ``look up'' (approximately
recover) the value of $\vek{v}_X[i]$, we compute the quantity
\[ \textit{CM}(i,\vek{b}_X) \equiv
    \min_{j=1}^{N_D} \vek{b}_X[ j, h_j(i) ]
\]
In other words, we look up the hashed value associated with $i$ in
each of the $N_D$ primitive sketches, and take the minimum value. The ``look up'' of the count-min sketch provides the following probabilistic guarantee, as proved in Supplementary Material B.

\begin{theorem}
Let $\vek{b}_X$ be a count-min sketch for $X$ of depth $N_D$ and
with $N_W > 2|X|$, and let $C \supseteq X$.
If $N_D > \log_2\frac{|C|}{\delta}$ then with probability at least 1-$\delta$,
$X$ can be recovered from $\vek{b}_X$ using $C$. 
\end{theorem}


To reconstruct a set from this encoding, we
first take the $k$ elements with highest dot product
$\vek{a}_X^{T}\vek{e}_i$, where $k$ is a fixed hyperparameter. This is
done efficiently with a maximum inner product search
\cite{mussmann2016learning} (MIPS), which we write
$\textnormal{TOP}_k(\textbf{a}_X,\vek{E})$.\footnote{ While
  $\textbf{a}_X$ could be based on other geometric representations for
  sets, we use MIPS queries because obtaining candidates this way can
  be very efficient \cite{mussmann2016learning}.}  These top $k$
elements are then filtered by the count-min sketch, resulting in a
sparse (no more than $k$ non-zeros) \emph{decoding} of the set
representation
\begin{equation} \nonumber 
   \hat{\vek{v}}_X[i] = 
     \left\{ \begin{array}{l}
        \textit{CM}(i, \vek{b}_X) \cdot \textnormal{softmax}(\vek{a}_X^T~\vek{e}_i)
            \mbox{~if~$i \in \textnormal{TOP}_k(\vek{a}_X, \vek{E})$} \\
         0  \mbox{~else}
        \end{array}
     \right.
\end{equation}

The two pairs of the centroid-sketch representation are complementary.
The region around a centroid will usually contain entities with many
similar properties, for example ``US mid-size cities,'' or ``Ph.D. students in
NLP'': conceptually, it can be viewed as defining 
a \emph{soft type} for the
entities in $X$.  However, simple geometric representations like centroids
are not expressive enough to encode arbitrary sets $X$, like ``Ph.D. students
presenting papers in session $z$''.  Count-min sketches do allow 
arbitrary weights to be stored, but may return incorrect values (with low probability) when queries.  However, in this scheme the sketch is only queried for $k$ candidates
close to the centroid, so it is possible to obtain very low 
error probabilities 
with small sketches
(discussed later).  

The centroid-sketch representation does assume that all elements of the
same set are similar in the sense that they all have the same ``soft
type''---i.e., are all in a sphere around a specific centroid.
It also assumes that sets are of size no more than $k$. 
(Note the radius of the
sphere is not learned---instead $k$ is simply a hyperparameter.)

\textbf{Faithfulness.} Below we define compositional operations
(like union, intersection, etc) on centroid-sketch set representations.
A representation produced this way is associated with a particular
logical definition of a set $X$ (e.g., $X=Z_1\cup Z_2$), 
and we say that the representation
is \emph{faithful} to that definition to the extent that it yields the appropriate
elements when decoded (which can be measured experimentally).

Experimentally sketches improve the faithfulness of \emql{}.  However,
\emph{the sketch part of a set representation is
  optional}---specifically it can be replaced with a vacuous sketch
that returns a weight of 1.0 whenever it is queried.\footnote{For 
  count-min sketches, if $\vek{b}_\mathbb{I}$ is an all-ones matrix of the
  correct size, then $\forall i~ \textit{CM}(i,\vek{b}_\mathbb{I})=1$.} Removing sketches is useful when one is
  focusing on generalization.

\textbf{Intersection and union}.  \label{sec:intersection} Set
interesection and union of sets $A$ and $B$ will be denoted as
$(\vek{a}_{A \cap B},\vek{b}_{A \cap B})$ and $(\vek{a}_{A \cup
  B},\vek{b}_{A \cup B})$, respectively. Both operations assume
that the soft types of $A$ and $B$ are similar, so we can
define the new centroids as
$$
\vek{a}_{A \cap B} = \vek{a}_{A \cup B} = \frac{1}{2}(\vek{a}_A + \vek{a}_B)
$$ 
\noindent To combine the sketches, we exploit the property (see Supplemental
Materials) that if $\vek{b}_A$ and $\vek{b}_B$ are sketches for $A$
and $B$ respectively, then a sketch for $A \cup B$ is
$\vek{b}_A+\vek{b}_B$, and the sketch for $A \cap B$ is
$\vek{b}_A\odot\vek{b}_B$ (where $\odot$ is Hadamard product).  Hence
\begin{align*}
\vek{b}_{A \cap B} =  \vek{b}_A \odot \vek{b}_B ~~~~~ \vek{b}_{A \cup B} =  \vek{b}_A + \vek{b}_B
\end{align*}


\textbf{Relational following.}
As noted above,
relation following takes a set of entities $X$ and a set of
relations $R$ and computes the set of entities related to something in
$X$ via some relation in $R$:
\begin{equation*}
X.\textit{follow}(R) \equiv
\{ y ~|~ \exists r\in{}R, x\in{}X:r(x,y)\}
\end{equation*}
where ``$r(x,y)$'' indicates that this triple is in the KB (other
notation is listed in Supplemental Material A.)  For example, to look up
the headquarters of the Apple company one might compute $Y =
X.\textit{follow}(R)$ where $X$ and $R$ are singleton sets containing
``\textit{Apple\_Inc}'' and ``\textit{headquarters\_of}''
respectively, and result set $Y = \{\textit{Cupertino}\}$. 

\label{sec:follow}
Relation following is implemented using an embedding matrix $\vek{K}$ for KB triples that parallels
the element embedding matrix $\vek{E}$: for every triple $t=r(x,y)$ in
the KB, $\vek{K}$ contains a row $\vek{r}_t=\left[\vek{e}_r;
  \vek{e}_x; \vek{e}_y\right]$ concatenating the embeddings for $r$,
$x$, and $y$.  To compute $Y=X.\textit{follow}(R)$ first we create a
query $\vek{q}_{R,X}=\left[\lambda \cdot \vek{a}_R; \vek{a}_X;
  \vek{0}\right]$ by concatenating the centroids for $R$ and $X$ and
padding it to the same dimension as the triple embeddings (and $\lambda$ is a
hyper-parameter scaling the weight of the relation).  Next using
the query $\textbf{q}_{R,X}$, we perform a MIPS search against all
triples in KB $\textbf{K}$ to get the top $k$ triples matching this
query, and these triples are re-scored with the sketches of $X$ and
$R$.  Let $\vek{r}_t = \left[\vek{e}_{r_i}; \vek{e}_{x_j};
  \vek{e}_{y_\ell}\right]$ be the representation of retrieved triple
$t=r_i(x_j,y_\ell)$. Its score is
$$ s(\vek{r}_t) = \textrm{CM}(i, \vek{b}_R) \cdot \textrm{CM}(j, \vek{b}_X) \cdot \textnormal{softmax}(\vek{q}_{R, X}^T~\vek{r}_t)
$$
We can then project out the objects from the top $k$ triples as a sparse $k$-hot vector:
$$ \hat{\vek{v}}_Y(\ell) = \sum_{\vek{r}_t \in \textnormal{TOP}_k(\textbf{q}_{R, X}, \vek{K}), t=\_(\_,y_\ell)} s(\vek{r}_t)
$$
Finally $\hat{\vek{v}}_Y$ is converted to a set representation $(\vek{a}_Y, \vek{b}_Y)$, which represents the output of the operation, $Y=X.\textit{follow}(R)$. The triple store used for implementing \textit{follow} is thus a kind
of key-value memory network~\cite{kvmem}, augmented with a
sparse-dense filter in the form of a count-min sketch.

\textbf{Relational filtering.} 
Relational filtering, similar to an existential restriction in description logics, removes from $X$ those
entities that are not related to something in set $Y$ via some
relation in $R$:
$$
X.\textit{filter}(R,Y) \equiv \{ x \in X | \exists r\in{}R, y\in{}Y:r(x,y)\}
$$
\noindent For example, $X.\textit{filter}(R,Y)$ would filter out the
companies in $X$ whose headquarters are not in Cupertino, if $R$ and $Y$ are as
in the previous example. 
Relational filtering is implemented
similarly to \textit{follow}. For $X.\textit{filter}(R, Y)$, the query
must also be aware of the objects of the triples, since they should be
in the set $Y$.  The query vector is thus
\(
    \vek{q}_{R, X, Y} = \left[\lambda \cdot \vek{a}_R; \vek{a}_X; \vek{a}_Y\right]
\).
Again, we perform a retrieval using query $\vek{q}_{R, X, Y}$, but we filter with subject, relation, and object sketches $\vek{b}_R$, $\vek{b}_X$, $\vek{b}_Y$, so the score of an encoded triple $\vek{r}_t$ is
\begin{align*} 
s(\vek{r}_t) &= \textrm{CM}(i, \vek{b}_R) \cdot  \textrm{CM}(j, \vek{b}_X) \cdot \textrm{CM}(\ell, \vek{b}_Y) \cdot  \textnormal{softmax}(\vek{q}_{R, X, Y}^T~\vek{r}_t)
\end{align*}
The same aggregation strategy is used as for the  \textit{follow} operation, except that scores are aggregated over the subject entities instead of objects.

\label{sec:union-discussion}
\textbf{Unions in \emql{} vs Query2Box.} By construction, all \emql{} operations
are closed under composition, because they all take and return the
same sparse-dense representations, and the computation graph
constructed from an \emql{} expression is similar in size and structure
to the original \emql{} expression.  We note this differs from Query2Box,
where union is implemented by rewriting a query into a normal form.  A disadvantage of the Query2Box normal-form approach is that the normal form 
can be exponentially larger than the original expression.  

However, a
disadvantage of \emql{}'s approach is that unions are only allowed between sets of similar ``soft types''.  In fact, \emql{}'s centroid-sketch representation will not compactly encode \emph{any} set of sufficiently diverse entities: in a small embedding space, a diverse set like $X=\{\textit{kangaroo}, \textit{ashtray}\}$ will have a centroid far from any element, so a top-$k$ MIPS query with small $k$ would have low recall.
This limitation of \emql{} could be addressed by introducing a second normal-form disjunction operation that outputs a union of centroid-sketch representations, much as Query2Box's disjunction outputs a union of boxes---however, we leave such an extension as a topic for future work.

\textbf{Size and density of sketches.} \label{sec:prop} Although the centroid-based
geometric constraints are not especially expressive, we note that
\emql{}'s sparse-dense representation can still express sets accurately, as
long as the $k$-nearest neighbor retrieval has good recall.
Concretely, consider a set $A$ with $\abs{A}=m$ and sparse-dense
representation $(\vek{a}_A,\vek{b}_A)$. Suppose that $k=cm$ ensures
that all $m$ elements of $A$ are retrieved as $k$-nearest neighbors of
$\vek{a}_A$; in other words, retrieval precision may be as low as
$1/c$. By Theorem~2 in the Supplementary Materials, a sketch of size
$2m\log_2\frac{cm}{\delta}$ will recover \emph{all} the weights in $A$
with probability at least $1-\delta$.

In our experiments we assume sets are of size $m<100$, and that
$c=10$.  Using 32 numbers per potential set member leads to $\delta
\approx \frac{1}{50}$ and a sketch size of about 4k.  Put another way,
sets of 100 elements require about as much storage as the BERT~\cite{devlin-etal-2019-bert}
contextual encoding of 4 tokens; alternatively the sketch for 100 elements
requires about 1/4 the storage of 100 embeddings with
$d=128$.\footnote{Of course, directly storing 100 embeddings is less
  useful for modeling, since that representation does not support
  operations like relation following or intersection.}

It is also easy to see that for a set of size $m$, close to half of
the numbers in the sketch will have non-zero values.  Thus only a
moderate savings in space is obtained by using a sparse-matrix data
structure: it is quite practical to encode sketches with GPU-friendly
dense-tensor data structures.

\textbf{Loss function.}  This representation requires entities that 
will be grouped into sets to be close in embedding space, so entity embeddings must
be trained to have this property---ideally, for all sets that arise in the
course of evaluating queries.  In the training process we use to encourage this property, an
example is a query (such as
``\textit{$\{$Apple\_Inc$\}$.follow($\{$headquarters\_of$\}$ $\cup$
  $\{$Sunnyvale$\}$}'') and a target output set $Y$.
Evaluation of the query produces an approximation $\hat{Y}$, encoded
as $(\hat{\vek{a}}_Y,\hat{\vek{b}}_Y)$, and the goal of training is
make $\hat{Y}$ approximate $Y$.

Let $\vek{v}_Y$ be the canonical $k$-hot encoding of $Y$.  While the
sketches prevent an element $y'\not\in{}\hat{Y}$ from getting too high
a score, the top-$k$ operator used to retrieve candidates only has high
recall if the elements in $\hat{Y}$ are close in the inner product
space.  We thus train embeddings to minimize
\[ \textnormal{cross\_entropy}(\textnormal{softmax}(\hat{\vek{a}_Y}^T, \vek{E}),
               \vek{v}_Y / |\!|\vek{v}_Y|\!|_1)
\]
Note that this objective ignores the sketch\footnote{The sketch is
not used for this objective, but is used in \S~\ref{sec:exp_kbe} where we train a QA system which includes \emql{} as a component. Hence in general it is necessary for inference with the sketch to be differentiable.}, so it forces the dense
representation to do the best job possible on its own. In training
$\hat{Y}$ can be primitive set, or the result of a computation (see \S~\ref{sec:exp_kbe}).

\section{Experiments} \label{sec:expts}
We evaluate \emql{} first intrinsically for its ability to model set expressions ~\cite{hamilton2018embedding}, and then extrinsically as the reasoning component in two multi-hop KB question answering benchmarks (KBQA). 


\subsection{Learning to reason with a KB \label{sec:exp_kbe}}

\textbf{Generalization.} To evaluate performance in generalizing to plausible answers, we follow the procedure of Ren et al. \cite{ren2020query2box} who considered nine different types of queries, as summarized in Table~\ref{tab:entailments}, 
and data automatically constructed from three widely used KB completion (KBC) benchmarks. Briefly, to evaluate performance for QE, Ren et al.\ first hold out some triples from the KB for validation and test, and take the remaining triples as the \emph{training KB}. Queries are generated randomly using the query templates of Table~\ref{tab:entailments}.  The gold answer for a query is the traditional logical evaluation on the \emph{full KB}, but the QE system is trained to approximate the gold answer using only the smaller \emph{training KB}.  Queries used to
evaluate the system are also constrained to \emph{not} be fully answerable using only logical entailment over the training KB.  
For details, see Supplementary Materials D.

\begin{figure}[h!]
\small
\centering
\begin{floatrow}
\capbtabbox{%
    \small
    \begin{tabular}{ll|ll} 
        \hline
          &  Query Template &  & Query Template \\ \hline
         1p & $X.\textit{follow}(R)$ & 
           ip & $(X_1.\textit{follow}(R_1) \cap X_2.\textit{follow}(R_2)).\textit{follow}(R)$ \\
        2p & $X.\textit{follow}(R_1).\textit{follow}(R_2)$ & 
           pi & $X_1.\textit{follow}(R_1).\textit{follow}(R_2) \cap X_2.\textit{follow}(R_3)$\\
        3p & $X.\textit{follow}(R_1).\textit{follow}(R_2).\textit{follow}(R_3)$ & 
           2u & $X_1.\textit{follow}(R_1) \cup X_2.\textit{follow}(R_2)$\\
        2i & $X_1.\textit{follow}(R_1) \cap X_2.\textit{follow}(R_2)$ &
           up & $(X_1.\textit{follow}(R_1) \cup X_2.\textit{follow}(R_2)).\textit{follow}(R)$ \\
        3i & $X_1.\textit{follow}(R_1) \cap X_2.\textit{follow}(R_2) \cap X_3.\textit{follow}(R_3)$ & & \\
        \hline
    \end{tabular}
}
{%
  \caption{\small Nine query templates used.  Query2Box is trained on templates 1p, 2p, 3p, 2i, and 3i. \emql{} is trained on a variation of 1p and set intersection. \label{tab:entailments}}%
}
\end{floatrow}
\end{figure}

Query2Box is trained on examples from only five reasoning tasks (1p, 2p, 3p, 2i, 3i), with the remainder held out to measure the ability to generalize to new query templates.\footnote{Of ccourse, test queries are always distinct from training queries.}  \emql{} was trained on only two tasks: \textit{relational following} (a variation of 1p), and \textit{set intersection}.  Specifically we define a ``basic set'' $X$ to be any set of entities that share the same property $y$ with relation $r$, i.e. $X = \{x | r(x, y)\}$. In training \emql{} to answer intersection queries $(X_1 \cap X_2)$,  we let $X_1$ and $X_2$ be non-disjoint basic sets, and for relational following (1p),
$X$ is a basic set and $R$ a singleton relation set.  Training, using the method proposed in \S \ref{sec:method}, produces entity and relation embeddings, and queries are then executed by computing the \emql{} representations for each subexpression in turn.\footnote{In particular, intermediate \emql{} representations are never ``decoded'' by converting them to entity lists.} Since we are testing generalization, rather then entailment, we replace $\vek{b}_{\hat{Y}}$ with a vacuous all-ones count-min sketch in the final set representation for a query (but not intermediate ones).

We compare \emql{} to two baseline models: GQE \cite{hamilton2018embedding} and Query2Box (Q2B) \cite{ren2020query2box}. 
The numbers are shown in Table \ref{tbl:q2b_results}.  Following the Query2Box paper \cite{ren2020query2box} we use $d=400$ for their model and report Hits@3 (see Supplementary Materials D for other metrics).  For \emql{}, we use $d=64$, $k = 1000$, $N_W = 2000$ and  $N_D = 20$ throughout.  In this setting, our model is slightly worse than Query2Box for the 1p queries, much worse for the 2u queries, and consistently better on all the more complex queries.  \emql{}'s difficulties with the 2u queries are because of its different approach to implementing union, in particular the kangaroo-ashtray problem discussed in \S\ref{sec:union-discussion}.

\begin{figure}[t]
\small
\centering
\begin{floatrow}
\capbtabbox{%
    \small
\begin{tabular}{l|ccccc|cccc|c||ccc}
\hline
   Generalization & \multicolumn{10}{c||}{Generalization on FB15k-237} & FB15k & NELL & FB15k-237\\
\hline
  & 1p   & 2p   & 3p   & 2i   & 3i   & ip   & pi   & 2u   & up  & Avg & Avg & Avg & Avg \\ \hline
   \hline
 GQE             & 40.5 & 21.3 & 15.5 & 29.8 & 41.1 & 8.5  & 18.2 & 16.9 & 16.3 & 23.1 & 38.7 & 24.8 & 23.1 \\
Q2B             & \textbf{46.7} & 24   & 18.6 & 32.4 & 45.3 & 10.8 & 20.5 & \textbf{23.9} & 19.3 & 26.8 & 48.4 & 30.6 & 26.8 \\
 ~~$+d$=2000            & 37.2  & 20.7  & 19.4  & 22.6  & 37.1  & 9.7  & 16.8  & 20.0  & 17.8 & 22.4 & 34.5 & 23.4 & 22.4 \\
 \emql{} (ours)     & 37.7 & \textbf{34.9} & \textbf{34.3} & \textbf{44.3} & \textbf{49.4} & \textbf{40.8} & \textbf{42.3} & 8.7  & 28.2 & \textbf{35.8} & \textbf{49.5} & \textbf{46.8} & \textbf{35.8}\\
 ~~$-$ sketch & 43.1 & 34.6 & 33.7 & 41.0 & 45.5 & 36.7 & 37.2 & 15.3 & \textbf{32.5} & 35.5 & 48.6 & \textbf{46.8} & 35.5\\
\hline
   Entailment & \multicolumn{10}{c||}{Entailment on FB15k-237} & FB15k & NELL & FB15k-237\\
   \hline
 Q2B             & 58.5 & 34.3 & 28.1 & 44.7 & 62.1 & 11.7 & 23.9 & 40.5 & 22.0 & 36.2 & 43.7 & 51.1 & 36.2 \\
~~+$d$=2000            & 50.7 & 30.1 & 26.1 & 34.8 & 55.2 & 11.4 & 20.6 & 32.8 & 21.5 & 31.5 & 38.3 & 43.7 & 31.5 \\
 \emql{} (ours)          & \textbf{100.0}  & \textbf{99.5} & \textbf{94.7} & \textbf{92.2} & \textbf{88.8} & \textbf{91.5} & \textbf{93.0} & \textbf{94.7} & \textbf{93.7} & \textbf{94.2} & \textbf{91.4} & \textbf{98.8} & \textbf{94.2} \\ 
~~$-$ sketch & 89.3 & 55.7 & 39.9 & 62.9 & 63.9 & 51.9 & 54.7 & 53.8 & 44.7 & 57.4 & 55.5 & 82.5 & 57.4 \\

 \hline
\end{tabular}
}
{%
  \caption{\small Hits@3 results on the Query2Box datasets. Please see Supplementary Materials for full results on FB15k and NELL995 datasets and for mean reciprocal rank results. \label{tbl:q2b_results} }%
}
\end{floatrow}
\end{figure}

\textbf{Entailment.} To test the ability to infer logically entailed answers, \emql{} and Q2B were trained with the full KB instead of the training KB, so only reasoning (not generalization) is required to find answers. As we argue above, it is important for a query language to be also be faithful to the KB when answering compositional logical queries. The results in Table \ref{tbl:q2b_results} show \emql{} dramatically outperforms Q2B on all tasks in this setting, with average Hits@3 raised from 36-51 to the 90's.  

To see if larger embeddings would improve Q2B's performance on entailment tasks, we increased the dimension size to $d=2000$, and observed a decrease in performance, relative  to the tuned value $d=400$~\cite{ren2020query2box}.\footnote{Here $d=2000$ was the largest value of $d$ supported by our GPUs. Note this is still much smaller than the number of entities, which would be number required to guarantee arbitrary sets could be memorized with boxes. .}
In the ablation experiment(\emql{}$-$sketch), we remove the sketch and only use the centroid to make predictions. The results are comparable for generalization, but worse for entailment.

\subsection{Question answering} \label{sec:expt_kbe}
To evaluate QE as a neural component in a larger system, we followed \cite{cohen2020scalable} and 
embed \emql{} as a reasoning component in a KBQA model.  The reasoner of  \cite{cohen2020scalable},
ReifKB, is a sparse-matrix ``reified KB''
rather than a QE method, which does not generalize, but is perfectly faithful for entailment questions. In this section we evaluate replacing ReifKB with \emql{}.

ReifKB was evaluated on 
two KBQA datasets, MetaQA~\cite{zhang2017variational} and
WebQuestionsSP~\cite{webqsp}, which access different KBs.
For both datasets, the input to the KBQA system is a
question $q$ in natural language and a set of entities $X_q$ mentioned in
the question, and the output is a set of answers $Y$  (but no information is given about the latent logical query that produces $Y$.)
\emql{}'s set operations were pre-trained for each KB as in \S~\ref{sec:exp_kbe},
and then the KB embeddings were fixed while training the remaining parts of the QA model. Please see the Supplementary Materials for details.

\textbf{The MetaQA model}. The MetaQA datasets~\cite{zhang2017variational} contain multi-hop
questions in the movie domain that are answerable using the WikiMovies KB~\cite{miller2016key},
e.g., ``When were the movies
directed by Christopher Nolan released?'').  One dataset (here called MetaQA2) has
300k 2-hop questions and the other (MetaQA3) has 300k 3-hop questions.
We used similar models as those used for ReifKB. The model\footnote{Here
$A - B$ is a non-compositional \textit{set difference} operator, see Supplementary Materials for details.} for 2-hop questions is given on the left of Table \ref{tbl:qa_models}, where $W_1$ and $W_2$ are learned parameters, $\vek{b}_\mathbb{I}$ is a vacuous sketch,
and \textit{encode}$(q)$ 
is obtained by pooling the (non-contextual) embeddings of words in $q$.
The 3-hop case is analogous.\footnote{An important difference is that in ReifKB $R_1$ and $R_2$ are $k$-hot representations of sets of relation ids, not centroids in embedding space.}



\begin{table}[h!]
    \small
    \centering
    \caption{\small \emql{} models for MetaQA and WebQuestionsSP datasets. \label{tbl:qa_models}}
    
    \begin{tabular}{l|l}
         \textbf{MetaQA} & \textbf{WebQuestionsSP} \\ \hline
        $\hat{Y} =  X_q.follow(R_1).follow(R_2) - X_q$ & $X_1 = X_q.follow(R_1^e)$ \\
        $R_1 =  (\vek{a}_1, \vek{b}_\mathbb{I}), ~ \vek{a}_1 = W_1^T \textit{encode}(q)$ & $X_2 = X_q.follow(R_1^{cvt}).follow(R_2^e)$ \\
        $R_2 =  (\vek{a}_2, \vek{b}_\mathbb{I}), ~ \vek{a}_2 = W_2^T \textit{encode}(q)$ & $\hat{Y} = X_1 \cup X_2 \cup (X_1 \cup X_2).filter(R_3, Z)$ \\\hline
    \end{tabular}
\end{table}

\textbf{The WebQuestionsSP model}. This dataset~\cite{webqsp}
contains 4,737 natural language questions generated from Freebase. 
Questions in WebQuestionsSP are a mixture of 1-hop and
2-hop questions, sometimes followed by a relational filtering operation, which are answerable using a subset of FreeBase.
The intermediate entities of 2-hop questions are always ``compound value type'' (CVT) entities---entities that do
not have names, but describe $n$-ary relationships between
entities. For example, the question ``Who is Barack Obama's
wife?'' might be answered with the query
\textit{
$X_q$.follow($R_1$).follow($R_2$).filter($R_3$,$Z$)}, where 
$X_q=\{\textit{Barack\_Obama}\}$, $R_1$,
$R_2$, and $R_3$ are the relations \textit{has\_marriage},
\textit{spouse}, and \textit{gender}, and $Z$ is the set
$\{\textit{female}\}$.  Here \textit{$X_q$.follow($R_1$)} produces a CVT node
representing a marriage relationship for a couple.
The model we use (see Table \ref{tbl:qa_models}, right) is similar to MetaQA model,
except that the final stage is a union of several submodels---namely,
chains of one and two follow operations, with or without relational
filtering.  The submodels for the $R$'s also similar to those for MetaQA,
except that we used a BERT~\cite{devlin-etal-2019-bert} encoding of the question, and augmented the entity embeddings with pre-trained BERT
representations of their surface forms (see Supplementary Materials.)

The model in ReifKB~\cite{cohen2020scalable} does not support relational filtering\footnote{The relational filtering operation is not defined for ReifKB, although it could be implemented with sequences of simpler operations.}, so for it we report results with the simpler model $\hat{Y} = X_1 \cup X_2$.  
Of course the \emql{} models also differ in being QE models, so they model relations as centroids in embedding space instead of $k$-hot vectors.

\begin{figure}
\small
\centering
\begin{floatrow}
\capbtabbox{%
    \small
    \begin{tabular}{l|c|c|c}
    \hline
              & MetaQA2 & MetaQA3 & WebQSP \\ \hline
    KV-Mem    & 82.7  & 48.9  & 46.7 \\ 
    GRAFT-Net & 94.8  & 77.7  & 70.3\\ 
    PullNet   & \textbf{99.9}  & 91.4  & 69.7\\ 
    EmbedKGQA & 98.8  & 94.8  & 66.6 \\
    ReifKB    & 81.1  & 72.3  & 52.7 \\ \hline
    \emql{} (ours)     & 98.6  & \textbf{99.1}  & \textbf{75.5}\\ 
    ~~$-$ filter  
                    & --  & -- & 65.2 \\
    ~~$-$ sketch    & 70.3   & 60.9  & 53.2\\
    \hline
    
    \end{tabular}
}{%
  \caption{\small Hits@1 of WebQuestionsSP, MetaQA2, and MetaQA3. GRAFT-Net and PullNet were re-run on Web\-Questions\-SP with oracle sets of question entities $X_q$.\label{tbl:metaqa}}%
}
\ffigbox{%
  \includegraphics[width=0.45\textwidth]{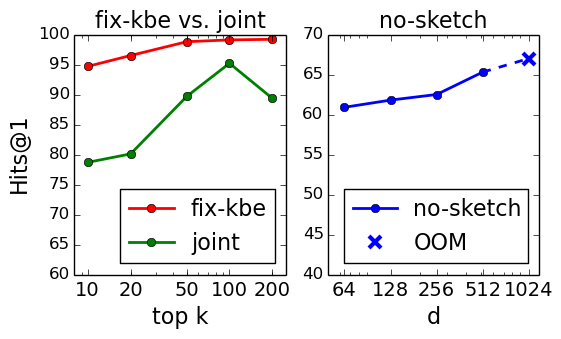}
}{%
  \caption{\footnotesize Hits@1 on MetaQA3. \textbf{Left:} Jointly training (joint) or fixed QE (fix-kbe) varying $k$ for top $k$ retrieval. \textbf{Right:} Varying 
  $d$ for \emqls{}. \label{tbl:wiki_ablation}}%
}
\end{floatrow}
\end{figure}

\textbf{Experimental results.} In addition to ReifKB~\cite{cohen2020scalable}, we report results for GRAPH-Net~\cite{graftnet} and PullNet~\cite{sun2019pullnet}, which are Graph-CNN based methods. EmbedKGQA~\cite{saxena2020improving} is the current state-of-the-art model that applies KB embeddings ComplEx~\cite{trouillon2016complex} in KBQA. Since our models make heavy use of the \textit{follow} operation, which is related to a
key-value memory, we also compare to a key-value memory network baseline~\cite{miller2016key}.
The results are shown on Table~\ref{tbl:metaqa}. 




On MetaQA3 and WebQSP datasets we exceed the previous state-of-the-art by a large margin (7.7\% and 5.8\% hits@1 absolute improvement), and results on MetaQA2 are comparable to the previous state-of-the-art. 
%
We also consider two ablated versions of our model, \emqls{} and \emqlf{}. \emqlf{} uses the same model as used in ReifKB~\cite{cohen2020scalable},  but still improves
over ReifKB significantly, showing the value of coupling learning with a QE system rather than a localist KB. \emqls{} disables sketches throughout (rather than only in the submodels for the $R$'s), and works consistently worse than the full model for all datasets. 
Thus it underscores the value of \emph{faithful QE} for KBQA tasks. (Notice that
errors in computing entailed answers will appear to the KBQA system as noise
in the end-to-end training process.)
Finally, Figure~\ref{tbl:wiki_ablation} (right) shows that performance of \emqls{} is
improved only slightly with much larger embeddings, also underscoring the value of the sketches.

As noted above, the QE component was first pre-trained on synthetic queries (as in \S~\ref{sec:exp_kbe}), and then the QA models were trained with fixed entity embeddings.  We also jointly trained the KB embeddings and the QA model for MetaQA3, using just the QA data. In this experiment, we also varied $k$, the top-$k$ entities retrieved at each step. Figure~\ref{tbl:wiki_ablation} (left) shows pre-training the KB embeddings (fix-kge) consistently outperforms jointly training KB embeddings for the QA model (joint), and demonstrates that pre-training QE on simple KB queries can be useful for downstream tasks.


\section{Concluding Discussion}


\emql{} is a new query embedding (QE) method, which combines a novel
centroid-sketch representation for entity sets with neural retrieval
over embedded KB triples.  In this paper we showed that \emql{} generalizes well, is differentiable, compact, scalable, and faithful with respect to deductive reasoning. However, there are areas for improvement.  Compared to the reified KB method~\cite{cohen2020scalable}, \emql{} learns and generalizes better, and does not rely on expensive sparse-matrix computations; however, unlike a reified KB, it requires KB-specific pretraining to find entity embeddings suitable for reasoning.  Like most
previous KBE or QE methods, \emql{} sets must correspond to neighborhoods in embedding space\footnote{A notable exception is Query2Box which represents sets with unions of rectangles using a non-compositional set union operator.  Although non-compositional set union could be added to \emql{} it is currently not implemented.}; \emql{}'s centroid-sketch representation additionally assumes that sets are of moderate cardinality.  Finally, in common with prior QE methods~\cite{hamilton2018embedding,ren2020query2box}, \emql{} does not support general logical negation, and has only very limited support for set difference. 

In spite of these limitations, we showed that
\emql{} substantially outperforms previous QE methods in
the usual experimental settings, and  massively outperforms
them with respect to faithfulness.  In addition to improving on the best-performing prior QE method, we demonstrated that it is possible to incorporate \emql{}
as a neural module in a KBQA system: to our knowledge this is the
first time that a QE system has been used to solve an downstream task
(i.e., a task other than KB completion).  Replacing a faithful
localist representation of a KB with \emql{} (but leaving rest of the QA
system intact) leads to double-digit improvements in Hits@1 on three benchmark tasks, and leads to a new state-of-the-art on the two more difficult tasks.

\newpage
\section*{Broader Impact}

\textbf{Overview.} This work addresses a general scientific question, query embedding (QE) for knowledge bases, and evaluates a new method, especially on a KB question-answering (KBQA) task.  A key notion in the work the \emph{faithfulness} of QE methods, that is, their agreement with deductive inference when the relevant premises are explicitly available. The main technical contribution of the paper is to show that massive improvements in faithfulness are possible, and that faithful QE systems can lead to substantial improvements in KBQA.  In the following, we discuss how these advances may affect risks and benefits of knowledge representation and question answering technology. 

\textbf{Query embedding.} QE, and more generally KBE, is a way of generalizing the contents of a KB by building a probabilistic model of the statements in, or entailed by, a KB.  This probabilistic model finds statements that could plausibly true, but are not explicitly stored: in essence it is a noisy classifier for possible facts. Two risks need to be considered in any deployment of such technology: first, the underlying KB may contain (mis)information that would improperly affect decisions; second, learned generalizations may be wrong or biased in a variety of ways that would lead to improperly justified decisions. In particular, training data might reflect societal biases that will be therebly incorporated into model predictions. Uses of these technologies should provide audit trails and recourse so that their predictions can be explained to and critiqued by affected parties.  

\textbf{KB question-answering.} General improvements to KBQA do not have a specific ethical burden, but like any other such technologies, their uses need to be subject to specific scrutiny. The general technology does require particular attention to accuracy-related risks. In particular, we propose a substantial ``softening'' of the typical KBQA architecture (which generally parses a question to produce a single hard KB query, rather than a soft mixture of embedded queries). In doing this we have replaced traditional KB, a mature and well-understood technology, with QE, a new and less well-understood technology.  Although our approach makes learning end-to-end from denotations more convenient, and helps us reach a new state-of-the-art on some benchmarks, it is  possible that replacing a hard queries to a KB with soft queries could lead to confusion as to whether answers arise from highly reliable KB facts, reliable reasoning over these facts, or are noise introduced by the soft QE system.  As in KBE/QE, this has consequences for downstream tasks is uncertain predictions are misinterpreted by users.

\textbf{Faithful QE.} By introducting the notion of faithfullness in studies of approximate knowledge representation in QE, we provided a conceptual yardstick for examining the accuracy and predictability of such systems. In particular, the centroid-sketch formalism we advocate often allows one to approximately distinguish entailed answers vs generalization-based answers by checking sketch membership.  In addition to quantitatively improving faithfulness, \emql{}'s set representation thus may qualitatively improve the interpretability of answers.  We leave further validation of this conjecture to future work.

\newpage

\bibliographystyle{abbrvnat}

\begin{thebibliography}{30}
\providecommand{\natexlab}[1]{#1}
\providecommand{\url}[1]{\texttt{#1}}
\expandafter\ifx\csname urlstyle\endcsname\relax
  \providecommand{\doi}[1]{doi: #1}\else
  \providecommand{\doi}{doi: \begingroup \urlstyle{rm}\Url}\fi

\bibitem[Bloom(1970)]{bloom1970space}
B.~H. Bloom.
\newblock Space/time trade-offs in hash coding with allowable errors.
\newblock \emph{Communications of the ACM}, 13\penalty0 (7):\penalty0 422--426,
  1970.

\bibitem[Bordes et~al.(2013)Bordes, Usunier, Garcia-Duran, Weston, and
  Yakhnenko]{bordes2013translating}
A.~Bordes, N.~Usunier, A.~Garcia-Duran, J.~Weston, and O.~Yakhnenko.
\newblock Translating embeddings for modeling multi-relational data.
\newblock In \emph{Advances in neural information processing systems}, pages
  2787--2795, 2013.

\bibitem[Cohen et~al.(2019)Cohen, Siegler, and Hofer]{nql2019}
W.~W. Cohen, M.~Siegler, and R.~A. Hofer.
\newblock Neural query language: {A} knowledge base query language for
  tensorflow.
\newblock \emph{CoRR}, abs/1905.06209, 2019.
\newblock URL \url{http://arxiv.org/abs/1905.06209}.

\bibitem[Cohen et~al.(2020)Cohen, Sun, Hofer, and Siegler]{cohen2020scalable}
W.~W. Cohen, H.~Sun, R.~A. Hofer, and M.~Siegler.
\newblock Scalable neural methods for reasoning with a symbolic knowledge base.
\newblock \emph{arXiv preprint arXiv:2002.06115}, 2020.
\newblock Appeared in ICLR-2020.

\bibitem[Cormode and Muthukrishnan(2005)]{cormode2005improved}
G.~Cormode and S.~Muthukrishnan.
\newblock An improved data stream summary: the count-min sketch and its
  applications.
\newblock \emph{Journal of Algorithms}, 55\penalty0 (1):\penalty0 58--75, 2005.

\bibitem[Daniely et~al.(2016)Daniely, Lazic, Singer, and
  Talwar]{daniely2016sketching}
A.~Daniely, N.~Lazic, Y.~Singer, and K.~Talwar.
\newblock Sketching and neural networks.
\newblock \emph{arXiv preprint arXiv:1604.05753}, 2016.

\bibitem[Das et~al.(2016)Das, Neelakantan, Belanger, and
  McCallum]{das2016chains}
R.~Das, A.~Neelakantan, D.~Belanger, and A.~McCallum.
\newblock Chains of reasoning over entities, relations, and text using
  recurrent neural networks.
\newblock \emph{arXiv preprint arXiv:1607.01426}, 2016.

\bibitem[Demeester et~al.(2016)Demeester, Rockt{\"a}schel, and
  Riedel]{demeester2016lifted}
T.~Demeester, T.~Rockt{\"a}schel, and S.~Riedel.
\newblock Lifted rule injection for relation embeddings.
\newblock \emph{arXiv preprint arXiv:1606.08359}, 2016.

\bibitem[Devlin et~al.(2018)Devlin, Chang, Lee, and Toutanova]{devlin2018bert}
J.~Devlin, M.-W. Chang, K.~Lee, and K.~Toutanova.
\newblock Bert: Pre-training of deep bidirectional transformers for language
  understanding.
\newblock \emph{arXiv preprint arXiv:1810.04805}, 2018.

\bibitem[Guu et~al.(2015)Guu, Miller, and Liang]{guu2015traversing}
K.~Guu, J.~Miller, and P.~Liang.
\newblock Traversing knowledge graphs in vector space.
\newblock \emph{arXiv preprint arXiv:1506.01094}, 2015.

\bibitem[Hamilton et~al.(2018)Hamilton, Bajaj, Zitnik, Jurafsky, and
  Leskovec]{hamilton2018embedding}
W.~Hamilton, P.~Bajaj, M.~Zitnik, D.~Jurafsky, and J.~Leskovec.
\newblock Embedding logical queries on knowledge graphs.
\newblock In \emph{Advances in Neural Information Processing Systems}, pages
  2026--2037, 2018.

\bibitem[Lin et~al.(2015)Lin, Liu, Luan, Sun, Rao, and Liu]{lin2015modeling}
Y.~Lin, Z.~Liu, H.~Luan, M.~Sun, S.~Rao, and S.~Liu.
\newblock Modeling relation paths for representation learning of knowledge
  bases.
\newblock \emph{arXiv preprint arXiv:1506.00379}, 2015.

\bibitem[Liu et~al.(2017)Liu, Wu, and Yang]{liu2017analogical}
H.~Liu, Y.~Wu, and Y.~Yang.
\newblock Analogical inference for multi-relational embeddings.
\newblock In \emph{Proceedings of the 34th International Conference on Machine
  Learning-Volume 70}, pages 2168--2178. JMLR. org, 2017.

\bibitem[Miller et~al.(2016{\natexlab{a}})Miller, Fisch, Dodge, Karimi, Bordes,
  and Weston]{miller2016key}
A.~Miller, A.~Fisch, J.~Dodge, A.-H. Karimi, A.~Bordes, and J.~Weston.
\newblock Key-value memory networks for directly reading documents.
\newblock \emph{EMNLP}, 2016{\natexlab{a}}.

\bibitem[Miller et~al.(2016{\natexlab{b}})Miller, Fisch, Dodge, Karimi, Bordes,
  and Weston]{kvmem}
A.~H. Miller, A.~Fisch, J.~Dodge, A.~Karimi, A.~Bordes, and J.~Weston.
\newblock Key-value memory networks for directly reading documents.
\newblock \emph{CoRR}, abs/1606.03126, 2016{\natexlab{b}}.
\newblock URL \url{http://arxiv.org/abs/1606.03126}.

\bibitem[Mussmann and Ermon(2016)]{mussmann2016learning}
S.~Mussmann and S.~Ermon.
\newblock Learning and inference via maximum inner product search.
\newblock In \emph{International Conference on Machine Learning}, pages
  2587--2596, 2016.

\bibitem[Rastogi et~al.(2017)Rastogi, Poliak, and
  Van~Durme]{rastogi2017training}
P.~Rastogi, A.~Poliak, and B.~Van~Durme.
\newblock Training relation embeddings under logical constraints.
\newblock In \emph{KG4IR@ SIGIR}, pages 25--31, 2017.

\bibitem[Ren et~al.(2020)Ren, Hu, and Leskovec]{ren2020query2box}
H.~Ren, W.~Hu, and J.~Leskovec.
\newblock Query2box: Reasoning over knowledge graphs in vector space using box
  embeddings.
\newblock \emph{arXiv preprint arXiv:2002.05969}, 2020.
\newblock Appeared in ICLR-2020.

\bibitem[Saxena et~al.(2020)Saxena, Tripathi, and
  Talukdar]{saxena2020improving}
A.~Saxena, A.~Tripathi, and P.~Talukdar.
\newblock Improving multi-hop question answering over knowledge graphs using
  knowledge base embeddings.
\newblock \emph{ACL}, 2020.

\bibitem[Sun et~al.(2018)Sun, Dhingra, Zaheer, Mazaitis, Salakhutdinov, and
  Cohen]{graftnet}
H.~Sun, B.~Dhingra, M.~Zaheer, K.~Mazaitis, R.~Salakhutdinov, and W.~W. Cohen.
\newblock Open domain question answering using early fusion of knowledge bases
  and text.
\newblock \emph{EMNLP}, 2018.

\bibitem[Sun et~al.(2019)Sun, Bedrax-Weiss, and Cohen]{sun2019pullnet}
H.~Sun, T.~Bedrax-Weiss, and W.~W. Cohen.
\newblock Pullnet: Open domain question answering with iterative retrieval on
  knowledge bases and text.
\newblock \emph{arXiv preprint arXiv:1904.09537}, 2019.

\bibitem[Svozil(1998)]{svozil1998quantum}
K.~Svozil.
\newblock \emph{Quantum logic}.
\newblock Springer Science \& Business Media, 1998.

\bibitem[Trouillon et~al.(2016)Trouillon, Welbl, Riedel, Gaussier, and
  Bouchard]{trouillon2016complex}
T.~Trouillon, J.~Welbl, S.~Riedel, {\'E}.~Gaussier, and G.~Bouchard.
\newblock Complex embeddings for simple link prediction.
\newblock In \emph{International Conference on Machine Learning}, pages
  2071--2080, 2016.

\bibitem[Vilnis and McCallum(2014)]{vilnis2014word}
L.~Vilnis and A.~McCallum.
\newblock Word representations via gaussian embedding.
\newblock \emph{arXiv preprint arXiv:1412.6623}, 2014.

\bibitem[Vilnis et~al.(2018)Vilnis, Li, Murty, and
  McCallum]{vilnis2018probabilistic}
L.~Vilnis, X.~Li, S.~Murty, and A.~McCallum.
\newblock Probabilistic embedding of knowledge graphs with box lattice
  measures.
\newblock \emph{arXiv preprint arXiv:1805.06627}, 2018.

\bibitem[Wang et~al.(2018)Wang, Wang, Liu, Chen, Zhang, and
  Qi]{emptysparql2018}
M.~Wang, R.~Wang, J.~Liu, Y.~Chen, L.~Zhang, and G.~Qi.
\newblock Towards empty answers in sparql: Approximating querying with rdf
  embedding.
\newblock In D.~Vrande{\v{c}}i{\'{c}}, K.~Bontcheva, M.~C. Su{\'a}rez-Figueroa,
  V.~Presutti, I.~Celino, M.~Sabou, L.-A. Kaffee, and E.~Simperl, editors,
  \emph{The Semantic Web -- ISWC 2018}, pages 513--529, Cham, 2018. Springer
  International Publishing.
\newblock ISBN 978-3-030-00671-6.

\bibitem[Wang et~al.(2017)Wang, Mao, Wang, and Guo]{wang2017knowledge}
Q.~Wang, Z.~Mao, B.~Wang, and L.~Guo.
\newblock Knowledge graph embedding: A survey of approaches and applications.
\newblock \emph{IEEE Transactions on Knowledge and Data Engineering},
  29\penalty0 (12):\penalty0 2724--2743, 2017.

\bibitem[Yih et~al.(2015)Yih, Chang, He, and Gao]{webqsp}
W.-t. Yih, M.-W. Chang, X.~He, and J.~Gao.
\newblock Semantic parsing via staged query graph generation: Question
  answering with knowledge base.
\newblock In \emph{Proceedings of the 53rd Annual Meeting of the Association
  for Computational Linguistics and the 7th International Joint Conference on
  Natural Language Processing (Volume 1: Long Papers)}, pages 1321--1331,
  Beijing, China, July 2015. Association for Computational Linguistics.
\newblock URL \url{http://www.aclweb.org/anthology/P15-1128}.

\bibitem[Zaheer et~al.(2017)Zaheer, Kottur, Ravanbakhsh, Poczos, Salakhutdinov,
  and Smola]{zaheer2017deep}
M.~Zaheer, S.~Kottur, S.~Ravanbakhsh, B.~Poczos, R.~R. Salakhutdinov, and A.~J.
  Smola.
\newblock Deep sets.
\newblock In \emph{Advances in neural information processing systems}, pages
  3391--3401, 2017.

\bibitem[Zhang et~al.(2018)Zhang, Dai, Kozareva, Smola, and
  Song]{zhang2017variational}
Y.~Zhang, H.~Dai, Z.~Kozareva, A.~J. Smola, and L.~Song.
\newblock Variational reasoning for question answering with knowledge graph.
\newblock In \emph{AAAI}, 2018.

\end{thebibliography}

\newpage
\appendix

\section{Notation} \label{sec:notation}
The notation used in this paper is summarized in Table~\ref{tab:notation}.
\begin{table*}[h]
\centerline{
\begin{tabular}{ll}
     $W,X,Y$ &  sets of entities \\
     $R$ & set of relations \\
     $r$ & a single relation \\
     $x,y$ & entities \\
     $x_i$ & entity with index $i$ \\
      & \\
     $A,B$ & set of anything (entities or relations) \\
     $U$ & universal set \\
     $\vek{v}_A$ & a $k$-hot vector for a set $A$\\
      & \\
     $r(x,y)$ & asserts this triple is in the KB\\
     $\vek{E}$ & matrix of entity embeddings \\
     $\vek{e}_x,\vek{e}_r$ & embedding of entity $x$, relation $r$ \\
     $\vek{e}_i$ & embedding of entity with index $i$,
        i.e. $\vek{e}_i = \vek{E}[i,:]$\\
     $\vek{KB}$ & matrix of triple embeddings,
        i.e., row for $r(x,y)$ is $[\vek{e}_r ; \vek{e}_x ; \vek{e}_y]$ \\
         & \\
     $(\vek{a}_X, \vek{b}_X)$ & area and sketch that represent set $X$ \\
     \textit{CM}$(i,\vek{b})$ & score for entity $i$ in the count-min sketch $\vek{b}$ \\
     $X.\textit{follow}(R)$ & soft version of 
        $\{ y ~|~ \exists r\in{}R, x\in{}X:r(x,y)\}$ \\
     $X.\textit{filter}(R,Y)$ & soft version of 
    $\{ x \in X ~|~ \exists r\in{}R, y\in{}Y:r(x,y)\}$ \\  
\end{tabular}}
\caption{Notation used in the paper, excluding notation used only in \S~\ref{app:sketch}}
\label{tab:notation}
\end{table*}

\section{Background on count-min sketches} \label{app:sketch}

\subsection{Definitions}

Count-min sketches \cite{cormode2005improved}
are a widely used randomized data structure.  We include this discussion for completeness, and our analysis largely follows \cite{daniely2016sketching}.

A count-min sketch, as used here, is an approximation of a vector
representation of a weighted set.
Assume a universe $U$ which is a set of integer ``object ids'' from
$\{1,\ldots,N\}$. A set $A \subseteq U$ can be encoded as a vector
$\vek{v}_A \in \R^n$ such that $\vek{v}_A[i]=0$ if $i\not\in S$, and
otherwise $\vek{v}_A[i]$ is a real-numbered weight for entity $i$ in
set $S$.  The purpose of the count-min sketch is to approximate
$\vek{v}_A$ with limited storage.

Let $h$ be a hash function mapping $\{1,\ldots,N\}$ to a smaller range
of integers $\{1,\ldots,N_W\}$, where $N_W \ll N$.  The
\emph{primitive sketch of $\vek{v}_A$ under $h$}, written
$\vek{s}_h(\vek{v}_A)$, is a vector such that
$$ \vek{s}_h(\vek{v}_A)[j] = \sum_{i:h(i)=j} \vek{v}_A[i]
$$ Algorithmically, this vector could be formed by starting with an
all-zero's vector of length $N_W$, then looping over every pair
$(i,w_i)$ where $w_i=\vek{v}_A[i]$ and incrementing each $\vek{s}_h[j]$ by $w_i$. 
A primitive sketch $\vek{s}_h$ contains some information about
$\vek{v}_A$: to look up the value $\vek{v}_A[i]$, we could look up
$\vek{s}_h[h(i)]$, and this will have the correct value if no other set
element $i'$ hashed to the same location.  We can improve this by
using multiple hash functions.

Specifically, let
$H=\{h_1,\ldots,h_{N_D}\}$ be a list of $N_D$ hash functions mapping
$\{1,\ldots,N\}$ to the smaller range of integers $\{1,\ldots,N_W\}$.
The \emph{count-min sketch $\vek{S}_H(\vek{v}_A)$ for a $\vek{v}_A$ under $H$} is a matrix such that each row $j$ is the primitive
sketch of $\vek{v}_A$ under $h_j$.  This sketch is an $N_W \times N_D$
matrix: $N_W$ is called the sketch width and $N_D$ is called the
sketch depth.

Let $\vek{S}$ be the count-min sketch for $A$.  To ``look up'' (approximately
recover) the value of $\vek{v}_A[i]$, we compute this quantity
\[ \textit{CM}(i,\vek{S}) \equiv
    \min_{j=1}^{N_D} \vek{S}[ j, h_j(i) ]
\]
In other words, we look up the hashed value associated with $i$ in
each of the $N_D$ primitive sketches, and take the minimum value.

\subsection{Linearity and implementation nodes}

Count-min sketches also have a useful ``linearity'' property, inherited from primitive sketches. It is easy to
show that for any two sets $A$ and $B$
represented by vectors $\vek{v}_A$ and $\vek{v}_B$
\begin{eqnarray*}
\vek{S}_H(\vek{v}_A +  \vek{v}_B) & = & \vek{S}_H(\vek{v}_A) +  \vek{S}_H(\vek{v}_B) \\
\vek{S}_H(\vek{v}_A \odot  \vek{v}_B) & = & \vek{S}_H(\vek{v}_A) \odot  \vek{S}_H(\vek{v}_B)
\end{eqnarray*}
Here, as elsewhere in this paper, $\odot$ is Hadamard product.


In general, although it is mathematically convenient to define the behavior of sketches in reference to $k$-hot vectors, it is \emph{not necessary to construct a vector $\vek{v}_A$ to construct a sketch}: all that is needed is the non-zero weights of the elements of $A$.  Alternatively, if one precomputes and stores the sketch for each singleton set, it is possible to create sketches for an arbitrary set by gathering and sum-pooling the sketches for each element.

\subsection{Probabilistic bounds on accuracy}

We assume the hash functions are random mappings from $\{1,\ldots,N\}$
to $\{1,\ldots,N_W\}$. More precisely, we assume that for all
$i\in\{1,\ldots,N\}$, and all $j\in\{1,\ldots,N_W\}$,
$ \Pr ( h_i(x)=a ) = \frac{1}{N_W}
$.
We will also assume that the $N_D$ hash functions are are all drawn
\emph{independently} at random. More precisely, for all $i\not= i'$,
$i,i' \in \{1,\ldots,N\}$, all $j,j'\in\{1,\ldots,N_D\}$ and all
$k,k'\in{}\{1,\ldots,N_W\}$,
$ \Pr ( h_j(i)=k \wedge  h_{j'}(i')=k' ) = \frac{1}{N_W^2} $.

Under this assumption, the probability of errors can be easily
bounded.  Suppose the sketch width is at least twice the cardinality
of $A$, i.e., $|A|<m$ and $N_W>2m$.  Then one can show for all
primitive sketches $j$:
$$ \Pr ( \vek{S}[ j,  h_j(i) ] \not= \vek{v}_A[i] ) \leq \frac{1}{2}                                                                                         
$$

From this one can show that the probability of any error in a
count-min sketch decreases exponentially in sketch depth.  (This result is a slight variant of one in \cite{daniely2016sketching}.)

\begin{theorem} \label{thm:lookup}
Assuming hash functions are random and independent as defined above,
then if $\vek{S}$ is a count-min sketch for $A$ of depth $N_D$, and
$N_W > 2|A|$, then
\[ \Pr(\textit{CM}(\vek{S},i) \not= \vek{v}_A[i]) ~\leq{}~ \frac{1}{2^{N_D}}
\]
\end{theorem}

This bound applies to a single CM operation.  However, by using a
union bound it is easy to assess the probability of making an error in
any of a series of CM operations.  In particular, we consider the
case that there is some set of candidates $C$ including all entities
in $A$, i.e., $A \subseteq C \subseteq U$, and consider recovering the
set $A$ by performing a CM lookup for every $i' \in C$.  Specifically, we
say that \emph{$A$ can be recovered from $\vek{S}$ using $C$} if $A                                                                                          
\subseteq C$ and
$$                                                                                                                                                           
\forall i' \in C, \textit{CM}(i', \vek{S}) = \vek{v}_A[i']                                                                                                
$$ Note that this implies the sketch must correctly score every $i'                                                                                          
\in C-A$ as zero. Applying the union bound to Theorem~\ref{thm:lookup}
leads to this result.

\begin{theorem} \label{thm:cand}
Let $\vek{S}$ be a count-min sketch for $A$ of depth $N_D$ and
with $N_W > 2|A|$, and let $C \supseteq A$.
If $N_D > \log_2\frac{|C|}{\delta}$ then with probability at least 1-$\delta$,
$A$ can be recovered from $\vek{S}$ using $C$.
\end{theorem}

Many other bounds are known for count-min sketches: perhaps the best-known result is that for $N_W > \frac{2}{\epsilon}$ and $N_D > \log \frac{1}{\delta}$, the probability that $\textit{CM}(i,\vek{S}) > \vek{v}_A[i] + \epsilon$ is no more than $\delta$ \cite{cormode2005improved}.  Because there are many reasonable formal bounds that might or might not apply in an experimental setting, typically the sketch shape is treated as a hyperparameter to be optimized in experimental settings.

\section{Set difference \label{app:set-diff}}
Another operation we use is set difference: e.g. ``movie directors but not writers''  requires one to compute a set difference $A_\textnormal{directors} - B_\textnormal{writers}$.  In computing a set difference, the soft-type of the output $A-B$ is the same as that of $A$, and we exclude the necessary elements from the count-min sketch to produce $(\vek{a}_{A - B}, \vek{b}_{A - B})$, where
\begin{align*}
\vek{a}_{A - B} &= \vek{a}_A\\
\vek{b}_{A - B} &= \vek{b}_A \odot (\vek{b} \neq 0)
\end{align*}
This is exact when $B$ is unweighted (the case we consider here), but only approximates set difference for general weighted sets.

\section{More experiment details}
\subsection{Learn to reason over a KB}\label{app:dataset_stats}
The statistics of the Query2Box datasets are shown in Table \ref{tbl:q2b_stats}.

\begin{table}[h!]
\centering
\begin{subtable}{\linewidth}
\centering
\begin{tabular}{c|cccccc}
\hline
          & Entities & Relations & Training Triples & Test Triples & Total Triples \\ \hline
FB15k     & 14,951   & 1,345     & 533,142       & 59,071       & 592,213       \\
FB15k-237 & 14,505   & 237       & 289,641       & 20,438       & 310,079       \\
NELL995   & 63,361   & 200       & 128,537       & 14,267       & 142,804 \\ \hline
\end{tabular}
\caption{Size of splits into train and test for all the Query2Box KBs.}
\end{subtable}
~\\~\\
\begin{subtable}{\linewidth}
\centering
\begin{tabular}{c|ccc|cc}
\hline
\multicolumn{1}{l}{} & \multicolumn{3}{|c|}{Train}               & \multicolumn{2}{c}{Test} \\ \hline
task                 & Basic sets & Follow (1p) & Intersection & Follow (1p)   & Others   \\ \hline
FB15k                & 11,611     & 96,750      & 355,966      & 67,016        & 8,000    \\
FB15k-237            & 11,243     & 50,711      & 191,934      & 22,812        & 5,000    \\
NELL995              & 19,112     & 36,469      & 108,958      & 17,034        & 4,000   \\ \hline
\end{tabular}
\caption{Number of training and testing examples of the Query2Box datasets. Training data for \emql{} are derived from the same training KB as Query2Box. \emql{} is directly evaluated on the same test data without further fine-tuning.}
\end{subtable}
\caption{Statistics for the Query2Box datasets. \label{tbl:q2b_stats}}
\end{table}

We also measure the MRR on the Query2Box datasets. The results are presented in Table \ref{tbl:q2b_hits3} and \ref{tbl:q2b_mrr}.
\begin{table}[h!]
\centering
\small
\begin{tabular}{cl|ccccc|cccc|c}
\hline
 \multicolumn{2}{c|}{\textit{generalization}}   & 1p   & 2p   & 3p   & 2i   & 3i   & ip   & pi   & 2u   & up  & Avg  \\ \hline
\multicolumn{1}{c|}{FB15k}     & GQE             & 63.6 & 34.6 & 25.0 & 51.5 & 62.4 & 15.1 & 31.0 & 37.6 & 27.3 & 38.7 \\
\multicolumn{1}{c|}{}          & Q2B             & \textbf{78.6} & 41.3 & 30.3 & 59.3 & 71.2 & 21.1 & 39.7 & \textbf{60.8} & 33.0 & 48.4 \\
\multicolumn{1}{c|}{}          & ~~$+d$=2000      & 54.3 & 32.0 & 27.0 & 35.5 & 50.7 & 13.7 & 27.0 & 44.1 & 26.3 & 34.5 \\
\multicolumn{1}{c|}{}          & \emql{}(ours)            & 42.4 & \textbf{50.2} & \textbf{45.9}& \textbf{63.7} & \textbf{70.0} & \textbf{60.7} & \textbf{61.4} & 9.0  & \textbf{42.6} & \textbf{49.5} \\ 
\multicolumn{1}{c|}{}          & ~~- sketch      & 50.6  & 46.7  & 41.6  & 61.8  & 67.3  & 54.2  & 53.5  & 21.6  & 40.0 & 48.6 \\ \hline
\multicolumn{1}{c|}{FB15k-237} & GQE             & 40.5 & 21.3 & 15.5 & 29.8 & 41.1 & 8.5  & 18.2 & 16.9 & 16.3 & 23.1 \\
\multicolumn{1}{c|}{}          & Q2B             & \textbf{46.7} & 24   & 18.6 & 32.4 & 45.3 & 10.8 & 20.5 & \textbf{23.9} & 19.3 & 26.8 \\
\multicolumn{1}{c|}{}          & ~~$+d$=2000         & 37.2  & 20.7  & 19.4  & 22.6  & 37.1  & 9.7  & 16.8  & 20.0  & 17.8 & 22.4 \\
\multicolumn{1}{c|}{}          & \emql{}(ours)            & 37.7 & \textbf{34.9} & \textbf{34.3} & \textbf{44.3} & \textbf{49.4} & \textbf{40.8} & \textbf{42.3} & 8.7  & 28.2 & \textbf{35.8} \\
\multicolumn{1}{c|}{}          & ~~- sketch      & 43.1 & 34.6 & 33.7 & 41.0 & 45.5 & 36.7 & 37.2 & 15.3 & \textbf{32.5} & 35.5 \\ \hline
\multicolumn{1}{c|}{NELL995}   & GQE             & 41.8 & 23.1 & 20.5 & 31.8 & 45.4 & 8.1  & 18.8 & 20.0 & 13.9 & 24.8 \\
\multicolumn{1}{c|}{}          & Q2B             & \textbf{55.5} & 26.6 & 23.3 & 34.3 & 48.0 & 13.2 & 21.2 & \textbf{36.9} & 16.3 & 30.6 \\
\multicolumn{1}{c|}{}          & ~~$+d$=2000         & 49.1 & 22.1 & 17.5 & 21.4 & 39.9 & 8.9 & 17.2 & 26.4 & 8.1 & 23.4\\
\multicolumn{1}{c|}{}          & \emql{}(ours)            & 41.5 & \textbf{40.4} & \textbf{38.6} & \textbf{62.9} & \textbf{74.5} & \textbf{49.8} & \textbf{64.8} & 12.6 & 35.8 & \textbf{46.8}\\
\multicolumn{1}{c|}{}          & ~~- sketch      & 48.3 & 39.5 & 35.2 & 57.2 & 69.0 & 48.0 & 59.9 & 25.9 & \textbf{38.2} & \textbf{46.8} \\ \hline
 \multicolumn{2}{c}{\textit{entailment}}  & &  & & & && & \\ \hline
\multicolumn{1}{c|}{FB15k}     & Q2B             & 68.0 & 39.4 & 32.7 & 48.5 & 65.3 & 16.2 & 32.9 & 61.4 & 28.9 & 43.7  \\
\multicolumn{1}{c|}{}          & ~~$+d$=2000         & 59.0 & 36.8 & 30.2 & 40.4 & 57.1 & 14.8 & 28.9 & 49.2 & 28.7 & 38.3  \\
\multicolumn{1}{c|}{}          & \emql{}(ours)           & \textbf{98.5} & \textbf{96.3} & \textbf{91.1} & \textbf{91.4} & \textbf{88.1} & \textbf{87.8} & \textbf{89.2} & \textbf{88.7} & \textbf{91.3} &  \textbf{91.4}\\ 
\multicolumn{1}{c|}{}          & ~~- sketch      & 85.1 & 50.8 & 42.4 & 64.4 & 66.1 & 50.4 & 53.8 & 43.2 & 42.7 & 55.5 \\ \hline 
\multicolumn{1}{c|}{FB15k-237} & Q2B             & 58.5 & 34.3 & 28.1 & 44.7 & 62.1 & 11.7 & 23.9 & 40.5 & 22.0 & 36.2 \\
\multicolumn{1}{c|}{}          & ~~$+d$=2000         & 50.7 & 30.1 & 26.1 & 34.8 & 55.2 & 11.4 & 20.6 & 32.8 & 21.5 & 31.5 \\
\multicolumn{1}{c|}{}          & \emql{}(ours)            & \textbf{100.0}  & \textbf{99.5} & \textbf{94.7} & \textbf{92.2} &
\textbf{88.8} & \textbf{91.5} & \textbf{93.0} & \textbf{94.7} & \textbf{93.7} & \textbf{94.2}\\ 
\multicolumn{1}{c|}{}          & ~~- sketch      & 89.3 & 55.7 & 39.9 & 62.9 & 63.9 & 51.9 & 54.7 & 53.8 & 44.7 & 57.4 \\ \hline
\multicolumn{1}{c|}{NELL995}   & Q2B             & 83.9 & 57.7 & 47.8 & 49.9 & 66.3 & 19.9 & 29.6 & 73.7 & 31.0 & 51.1 \\
\multicolumn{1}{c|}{}          & ~~$+d$=2000         & 75.7 & 49.9 & 36.9 & 40.5 & 60.1 & 17.1 & 25.6 & 63.5 & 24.4 & 43.7 \\
\multicolumn{1}{c|}{}          & \emql{}(ours)            & \textbf{99.0} & \textbf{99.0} & \textbf{97.1} & \textbf{99.7} & \textbf{99.6} & 
\textbf{98.7} & \textbf{98.9} & \textbf{98.8} & \textbf{98.5} & \textbf{98.8}\\
\multicolumn{1}{c|}{}          & ~~- sketch      & 94.5 & 77.4 & 52.9 & 97.4 & 97.5 & 88.1 & 90.8 & 70.4 & 73.5 & 82.5 \\ \hline
\end{tabular}
\caption{Detailed Hits@3 results for all the Query2Box datasets. \label{tbl:q2b_hits3}}
\end{table}

\begin{table}[h!]
\centering
\small
\begin{tabular}{cl|ccccc|cccc|c}
\hline
 \multicolumn{2}{c|}{\textit{generalization}}   & 1p   & 2p   & 3p   & 2i   & 3i   & ip   & pi   & 2u   & up  & Avg \\ \hline
\multicolumn{1}{c|}{FB15k}     & GQE             & 0.505 & 0.320 & 0.222 & 0.439 & 0.536 & 0.142 & 0.280 & 0.300 & 0.242 & 0.332 \\
\multicolumn{1}{c|}{}          & Q2B             & \textbf{0.654} & 0.373 & 0.274 & 0.488 & 0.602 & 0.194 & 0.339 & \textbf{0.468} & 0.301 & 0.410 \\
\multicolumn{1}{c|}{}          & ~~$+d$=2000         & 0.461 & 0.289 & 0.242 & 0.292 & 0.421 & 0.130 & 0.236 & 0.342 & 0.235 & 0.294 \\
\multicolumn{1}{c|}{}          & \emql{}(ours)            & 0.368 & \textbf{0.452} & \textbf{0.409} & \textbf{0.574} & \textbf{0.609} & \textbf{0.556} & \textbf{0.538} & 0.074 & \textbf{0.375} & \textbf{0.439}\\
\multicolumn{1}{c|}{}          & ~~- sketch      & 0.453 & 0.418 & 0.362 & 0.556 & 0.592 & 0.503 & 0.482 & 0.182 & 0.351 & 0.433 \\  \hline
\multicolumn{1}{c|}{FB15k-237} & GQE             & 0.346 & 0.193 & 0.145 & 0.250  & 0.355 & 0.086 & 0.156 & 0.145 & 0.151 & 0.203\\
\multicolumn{1}{c|}{}          & Q2B             & \textbf{0.400} & 0.225 & 0.173 & 0.275 & 0.378 & 0.105 & 0.18  & \textbf{0.198} & 0.178 & 0.235\\
\multicolumn{1}{c|}{}          & ~~$+d$=2000         & 0.322 & 0.196 & 0.185 & 0.193 & 0.318 & 0.095 & 0.149 & 0.174 & 0.166 & 0.200 \\
\multicolumn{1}{c|}{}          & \emql{}(ours)            & 0.334 & \textbf{0.305} & \textbf{0.304} & \textbf{0.378} & \textbf{0.436} & \textbf{0.351} & \textbf{0.358} & 0.075 & 0.241 & \textbf{0.309}\\
\multicolumn{1}{c|}{}          & ~~- sketch      & 0.370 & 0.297 & 0.306 & 0.345 & 0.400 & 0.311 & 0.306 & 0.129 &\textbf{0.272} & 0.304\\ \hline
\multicolumn{1}{c|}{NELL995}   & GQE             & 0.311 & 0.193 & 0.175 & 0.275 & 0.408 & 0.080 & 0.170 & 0.159 & 0.130 & 0.211 \\
\multicolumn{1}{c|}{}          & Q2B             & \textbf{0.413} & 0.227 & 0.208 & 0.288 & 0.414 & 0.125 & 0.193 & \textbf{0.266} & 0.155 & 0.254\\
\multicolumn{1}{c|}{}          & ~~$+d$=2000       & 0.308 & 0.174 & 0.151 & 0.171 & 0.350 & 0.083 & 0.150 & 0.183 & 0.087 & 0.184 \\
\multicolumn{1}{c|}{}          & \emql{}(ours)            & 0.372 & \textbf{0.351} & \textbf{0.349} & \textbf{0.539} & \textbf{0.654} & \textbf{0.441} & \textbf{0.561} & 0.105 & 0.311 & \textbf{0.409} \\
\multicolumn{1}{c|}{}          & ~~- sketch      & 0.431 & 0.349 & 0.300 & 0.493 & 0.588 & 0.423 & 0.527 & 0.22 & \textbf{0.324} & 0.406\\  \hline
 \multicolumn{2}{c}{\textit{entailment}}  & &  & & & && & \\ \hline
\multicolumn{1}{c|}{FB15k}     & Q2B             & 0.559 & 0.347 & 0.288 & 0.389 & 0.553 & 0.145 & 0.280 & 0.444 & 0.257 & 0.362 \\
\multicolumn{1}{c|}{}          & ~~$+d$=2000         & 0.498 & 0.327 & 0.274 & 0.336 & 0.492 & 0.139 & 0.251 & 0.386 & 0.257 & 0.329 \\
\multicolumn{1}{c|}{}          & \emql{}(ours)            & \textbf{0.983} & \textbf{0.961} & \textbf{0.908} & \textbf{0.908} & \textbf{0.872} & \textbf{0.881} & \textbf{0.883} & \textbf{0.887} & \textbf{0.910} & \textbf{0.910} \\ 
\multicolumn{1}{c|}{}          & ~~- sketch      & 0.819 & 0.448 & 0.368 & 0.564 & 0.580 & 0.420 & 0.466 & 0.385 & 0.383 & 0.492 \\ \hline
\multicolumn{1}{c|}{FB15k-237} & Q2B             & 0.476 & 0.301 & 0.249 & 0.364 & 0.638 & 0.113 & 0.207 & 0.311 & 0.203 & 0.318\\
\multicolumn{1}{c|}{}          & ~~$+d$=2000         & 0.432 & 0.262 & 0.233 & 0.292 & 0.466 & 0.109 & 0.183 & 0.255 & 0.198 & 0.270 \\
\multicolumn{1}{c|}{}          & \emql{}(ours)            & \textbf{0.998} & \textbf{0.988} & \textbf{0.949} & \textbf{0.902} & \textbf{0.867} & \textbf{0.892} & \textbf{0.909} & \textbf{0.947} & \textbf{0.934} & \textbf{0.932}\\ 
\multicolumn{1}{c|}{}          & ~~- sketch      & 0.861 & 0.504 & 0.352 & 0.554 & 0.581 & 0.451 & 0.475 & 0.499 & 0.400 & 0.520 \\ \hline
\multicolumn{1}{c|}{NELL995}   & Q2B             & 0.652 & 0.465 & 0.412 & 0.420 & 0.562 & 0.186 & 0.257 & 0.516 & 0.269 & 0.415\\
\multicolumn{1}{c|}{}          & ~~$+d$=2000         & 0.545 & 0.409 & 0.331 & 0.357 & 0.526 & 0.155 & 0.217 & 0.399 & 0.253 & 0.355\\
\multicolumn{1}{c|}{}          & \emql{}(ours)            & \textbf{0.990} & \textbf{0.990} & \textbf{0.971} & \textbf{0.996} & \textbf{0.996} & \textbf{0.987} & \textbf{0.987} & \textbf{0.988} & \textbf{0.985} & \textbf{0.988}\\ 
\multicolumn{1}{c|}{}          & ~~- sketch      & 0.939 & 0.750 & 0.462 & 0.952 & 0.954 & 0.851 & 0.871 & 0.653 & 0.702 & 0.793 \\ \hline
\end{tabular}
\caption{MRR results on the Query2Box datasets. \label{tbl:q2b_mrr}}
\end{table}

\subsection{Question answering \label{sec:more-results-webqsp}}

\subsubsection{Datasets}
The statistics of MetaQA and WebQuestionsSP datasets are listed in Table~\ref{tbl:data-stats}. For WebQuestionsSP, we used a subset of Freebase obtained by gathering triples that are within 2-hops of
the topic entities in Freebase. We exclude a few extremely common
entities and restrict our KB subset so there are at most 100 tail entities for
each subject/relation pair (reflecting the limitation of our model to sets of cardinality less than 100). 

\begin{table}[h!]
\centering
\begin{subtable}{\linewidth}
\centering
\begin{tabular}{lccc}
\hline
               & Train   & Dev    & Test   \\
\hline
MetaQA 2-hop   & 118,980 & 14,872 & 14,872 \\
MetaQA 3-hop   & 114,196 & 14,274 & 14,274 \\
WebQuestionsSP & 2,848   & 250    & 1,639  \\
\hline
\end{tabular}
\caption{Number of train/dev/test data}
\end{subtable}
~\\
\begin{subtable}{\linewidth}
\centering
\begin{tabular}{lccc}
\hline
               & Triples   & Entities    & Relations   \\
\hline
MetaQA   & 392,906 & 43,230 & 18 \\
WebQuestionsSP & 1,352,735   & 904,938    & 695  \\
\hline
\end{tabular}
\caption{Size of KB}
\end{subtable}
\caption{Statistics for the MetaQA and WebQuestionsSP datasets. \label{tbl:data-stats}}
\end{table}

\subsubsection{MetaQA \label{app:metaqa-details}}
MetaQA makes use of the set difference operation.  For example, to answer the question ``What are other movies that have the same director as \textit{Inception}?'', we need to first find the director of \textit{Inception}, \textit{Christopher Nolan}, and all movies directed by him. Since the question above asks about
\emph{other} movies, the model should also remove the movie
\textit{Inception} from this set to obtain the final answer set $Y$. Thus in the first line of our model, we write
\begin{align*}
\hat{Y} & =  X_q.follow(R_1).follow(R_2) - X_q 
\end{align*}

For MetaQA, the entity embedding is just a learned lookup table. The question representation $\textit{encode}(q)$ is computed with a bag-of-word approach, i.e., an average pooling on the word embeddings of question $q$. 
The embedding size is 64, and scaling parameter for relation $\lambda$ is 1.0. Our count-min sketch has depth $N_D=20$ and width $N_W=500$. We set $k=100$ to be the number of entities we retrieve at each step,  and we pre-train KB embeddings and fix the embeddings when training our QA model.

\subsubsection{WebQuestionsSP \label{app:webqsp-details}}
We use pre-trained BERT to encode our question $q$, i.e.,  $\textit{encode}(q)$ is the BERT embedding of the [CLS] token. The relation sets $R_1$, $R_2$, $R_3$ are linear projections of the question embedding $\textit{encode}(q)$ paired with a vacuous all-ones sketch $\vek{b}_\mathbb{I}$. Relation centroids are stacked with one extra dimension that encodes the hard-type of entities: here the hard-type is a binary value that indicates if the entity is a $cvt$ node or not. 

For this dataset, to make the entities and relations easier to predict
from language, the embedding of each entity was adapted to include
a transformation of the BERT encoding of the surface form of the
entity names. Let $\vek{e}^0_x$ be the embedding of the [CLS] token
from a BERT~\cite{devlin-etal-2019-bert} encoding of the canonical name for
entity $x$, and let $\vek{e}^1_x$ be a vector unique to $x$.  Our
pre-trained embedding for $x$ is then $\vek{e}_x
= \left[W^T \vek{e}^0_x ; \vek{e}^1_x \right] p$, where $W$ is a
learned projection matrix.  The embedding of relation $r$ is set to
the BERT encoding ([CLS] token) of the canonical name of relation $r$.
In this experiments the BERT embeddings are transformed to 128
dimensions and the entity-specific portion $\vek{e}^1_x$ has a
dimension of 32. The scaling parameter for relation $\lambda$ is
0.1. 

The KB embedding is fixed after pre-training. We use a count-min
sketch with depth $N_D=20$ and width $N_W=2000$, and we retrieve
$k=1000$ intermediate results at each step.


In the ablation study, we did two more experiments on the WebQuestionsSP dataset. First, we
remove the BERT pre-trained embedding, and instead randomly initialize
the KB entity and relation embeddings, and train the set
operations. The performance of \emql{} (no-bert) on the downstream QA
task is 1.3\% lower than our full model. Second, we replace the exact
MIPS with a fast maximal inner-product
search~\cite{mussmann2016learning}. This fast MIPS is an approximation
of MIPS that eventually causes 2.1\% drop in performance
(Table \ref{tbl:abla-webqsp}).
\begin{table}[htp]
\centering
\begin{tabular}{l|c}
\hline
          & WebQuestionsSP \\ \hline
\emql{}      & \textbf{75.5}\\ 
\emql{} (no-sketch)   & 53.2\\
\emql{} (no-filter)   & 65.2 \\ 
\emql{} (approx. MIPS) & 73.4\\ 
\emql{} (no-bert)  & 74.2\\ \hline

\end{tabular}
\caption{Ablation study on WebQuestionsSP\label{tbl:abla-webqsp}}
\end{table}

\end{document}